\documentclass[11pt]{article}

\pdfoutput=1

\usepackage[T1]{fontenc}
\usepackage{mlmodern}

\usepackage{amsthm}
\usepackage{amsfonts}
\usepackage{amssymb}
\usepackage{amsmath}
\usepackage{bm}

\usepackage{graphicx}
\usepackage{xcolor}

\theoremstyle{definition}
\newtheorem{definition}{Definition}[section]
\newtheorem{example}{Example}[section]
\newtheorem{proposition}{Proposition}[section]
\newtheorem{remark}{Remark}[section]
\newtheorem{lemma}{Lemma}[section]
\newtheorem{theorem}{Theorem}[section]

\title{\bf Constrained Exploration in Reinforcement Learning  with Optimality Preservation}

\author{Peter C.~Y.~Chen \\ 
       Department of Mechanical Engineering\\
       National University of Singapore\\
       Singapore 117576 \\
       {\footnotesize Email: {\sf mpechenp@nus.edu.sg}}
}
\date{}

\begin{document}

\maketitle

\begin{abstract}	 
We consider a class of  reinforcement-learning systems in which the agent follows a behavior policy to explore a discrete state-action space to find an optimal policy while adhering to some restriction on its  behavior.
 Such restriction  may prevent the agent from visiting some state-action pairs, possibly leading to the agent finding only a sub-optimal policy. To address this problem we introduce the concept of  constrained exploration with optimality preservation, whereby the exploration behavior of the agent is constrained to meet a specification while the optimality of the (original) unconstrained learning process is preserved.
 We first  establish a feedback-control structure  that models the dynamics of  the unconstrained learning process. We then extend this structure by adding a  supervisor to ensure that the behavior of the agent meets the specification, and establish  (for a class of reinforcement-learning problems with a known deterministic environment) a necessary and sufficient condition under which optimality is preserved. This work demonstrates the utility and the prospect of studying reinforcement-learning problems in the context of the theories of discrete-event systems, automata and formal languages.

\vspace*{0.2in}
\noindent
{\bf Keywords}
{\em Constrained exploration, desired behavior, supervision, optimality preservation, automata}
 
\end{abstract}

\newpage

\section{Introduction}

In reinforcement learning,  exploration refers to the agent  taking actions according to a behavior policy in order to  traverse a typically discrete state space and collect rewards.  While exploring the state space, the agent uses an update rule to estimate, based on the rewards collected, the $Q$-values (i.e., state-action values) from one iteration to the next. If the $Q$-values converge to their optimums, an optimal policy can then be obtained. 
 For a class of reinforcement learning problems, such convergence is guaranteed under the Robbins-Monro conditions \cite{robbins1951stochastic}.

A requirement for satisfying  the Robbins-Monro conditions is that every state-action pair must have a non-zero probability of being visited by the agent\,---\,also known as {\em persistent exploration}.
If we consider the agent taking an action (when it is at a state) as `generating' a symbol denoting  that action, the sequences of actions thus generated by the agent as it traverses through the states represent the behavior of the agent.
For an episodic learning process, the behavior of the agent consists of all possible  action sequences  from the initial state to the set of goal states.   We refer to such a process as an {\em unconstrained}  learning process, and the associated optimal $Q$-values as the {\em intrinsic optimums}.

\subsection{The constrained exploration problem}
\label{sec:the_problem}

It may be useful to require the agent to exhibit certain desired behavior during exploration.  The typical  motivation for imposing such requirements is to achieve better performance in a learning  process.  For example, we may require  the agent to avoid some states that may be considered unsafe, or some action sequences that are considered undesirable (such as  the case of a robot repeatedly touching the ball in robot soccer as described in \cite{ng_policy_1999}).

In practice, we might also need to impose certain requirements just to accommodate some operational characteristics of the agent itself.
 For instance, suppose that due to some technical glitch in its motor drive, a wheeled robot (i.e., the agent) is more likely to malfunction if it is commanded to take two or more consecutive right turns. To reduce the risk of robot failure, we might  impose the requirement that during exploration the robot is not allowed to take two right turns in succession.  Imposition of this type of requirements may not necessarily be motivated by the goal of achieving better learning performance; it could simply be a way to cope with certain (possibly unexpected) limitation of the operational capability of the agent.

To meet these requirements  necessitates a mechanism to  manipulate the behavior of the agent during the learning process. 
We refer to such a mechanism as a {\em supervisor}. 
We consider a supervisor as {\em effective} if it forces the agent
to stay within a prescribed subset of its unconstrained behavior. We call
the learning process in which the agent is under such manipulation  a {\em  constrained}   learning process. 

Since the behavior of the agent is  described by the sequences of actions taken by the agent as it traverses through the states, manipulating the behavior of the agent means that the agent may be prevented by the supervisor from taking certain actions  when it reaches certain states.
Such manipulation raises the possibility that  some state-action pairs will not be visited by the agent throughout the {\em entire} learning process, thus breaking persistent exploration. As a result, the $Q$-values may not converge to their  intrinsic optimums. The following example illustrates this issue.

\begin{example}
\label{exp:visitability} 
Figure \ref{fig:Visitability} illustrates  two cases of desired behavior of an agent traversing in a $4 \times 4$ grid environment.  In the first case, illustrated in Figure \ref{fig:Visitability}(a),  the agent is not allowed to take two consecutive {\sf  right} (i.e., $a_{2}$) actions , while in the second, Figure \ref{fig:Visitability}(b), the agent is allowed to take a {\sf  right} action {\em only} immediately after having taken two consecutive {\sf  up} (i.e., $a_{1}$) actions.
\begin{figure}[!h]\centering
\scalebox{0.6}[0.6]{\includegraphics{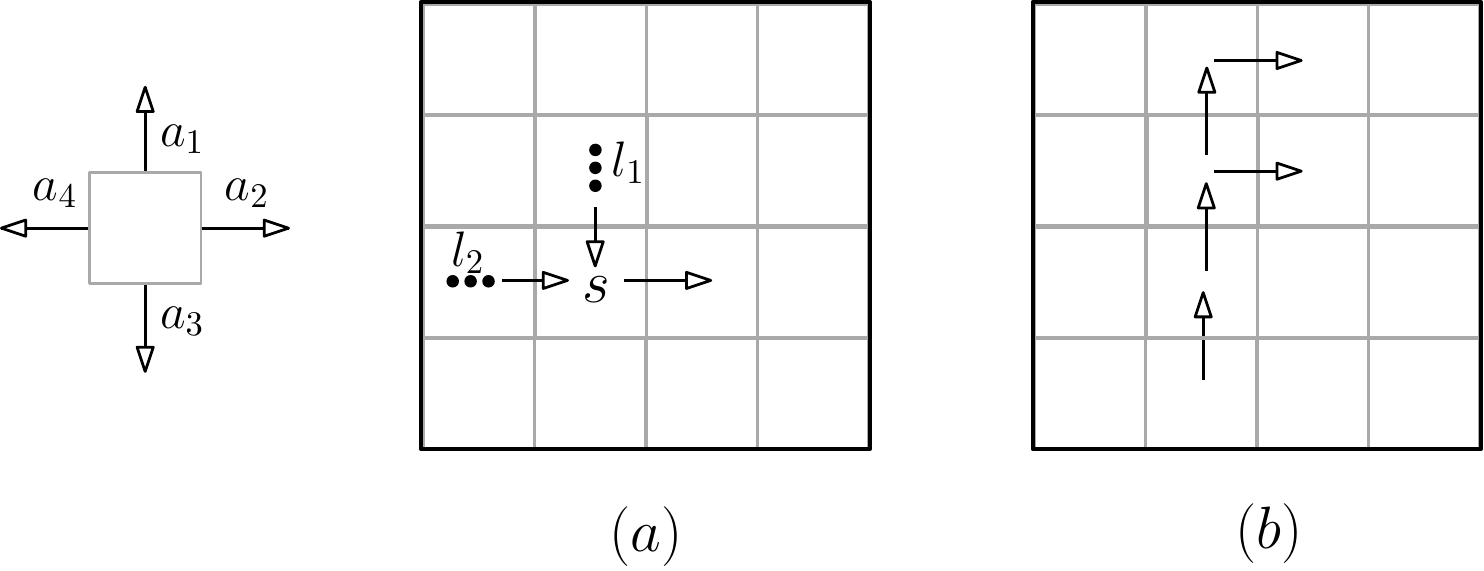}}
\caption{{\em Left}: The actions at a state. {\em (a)} and {\em (b)}: Two cases of desired behavior.}
\label{fig:Visitability}
\end{figure}

 Consider the situation in Figure \ref{fig:Visitability}(a). If the agent reaches $s$ by following the action sequence $l_{2}$, it is then prohibited from taking the {\sf right} action at $s$; however, it is permitted to do so if it reaches $s$ via $l_{1}$. Hence, all of the state-action pairs $(s,a_{i})$, where $i\in \{1,2,3,4\}$, have an non-zero probability of being visited by the agent.
For the situation  in Figure \ref{fig:Visitability}(b), the agent is allowed to take a {\sf  right} action {\em only} immediately after having taken two consecutive {\sf  up} (i.e., $a_{1}$) actions. This results in the agent being prevented from visiting all state-action pairs $(s, a_{2})$, where $s$ is any state in the lower two rows of the grid; hence, the probability of these state-action pairs being visited by the agent is zero.
\end{example}

Ideally we seek a supervisor that, while  manipulating the exploration behavior of the agent,  does not interfere with the $Q$-values converging to their intrinsic optimums. We refer to such a supervisor as {\em optimality-preserving}. 

We describe the {\em   constrained exploration problem} as follows. {\em For a given desired behavior of the agent, find a supervisor that is both effective and optimality-preserving}.

In this paper we investigate the  constrained exploration problem by first developing an automaton-model that describes the dynamics of the unconstrained  reinforcement learning process, then formulating a supervisory feedback-control structure incorporating this automaton-model. We next prove  (for a given desired behavior of the agent)  the existence of an effective supervisor and establish its realization in the form of an automaton  without any knowledge about  the dynamics of the environment.  We then determine a necessary and sufficient condition under which an effective supervisor is optimality-preserving  for a class of reinforcement-learning problems under the assumption that the dynamics of the  environment is deterministic and known.

\subsection{Literature survey}
\label{sec:literature}

In the standard setting of a reinforcement learning problem, the exploration behavior of the agent is determined by the behavior policy (e.g.,  $\epsilon$-greedy) that governs the action-selection process. By incorporating external knowledge into the action-selection process,  the exploration behavior of the agent can be manipulated to improve the speed of convergence or to meet certain operational requirements.
 
\subsubsection{Incorporating external knowledge for faster  convergence}

Incorporating external knowledge for the purpose of improving the speed of convergence enables policies to be learned based on additional information about the structure of the problem or based on expert intervention. 
Techniques that are based on additional information about the structure of the problem include reward shaping by modification of the reward function (e.g., \cite{ng_policy_1999, laud_influence_2003}), behavior cloning (e.g., \cite{nair_overcoming_2018}), direct policy learning (e.g., \cite{menda_ensembledagger_2019, ross_efficient_2010,zhang_query-efficient_2017}),   and inverse reinforcement learning (e.g., \cite{abbeel_apprenticeship_2004}).  
 Techniques for improving the speed of learning  based on expert intervention  include teacher-student framework (e.g., \cite{torrey2013teaching, amir_interactive_2016}), human-feedback for policy shaping (e.g., \cite{cederborg_policy_2015, griffith_policy_2013}),  transfer of learned actions or skill priors (e.g.,\cite{sherstov_improving_2005,pertsch_accelerating_2021}), and human-in-the-loop action removal or selection (e.g., \cite{abel_agent-agnostic_2017, rosman_what_2012, mandel_where_2017}).

\subsubsection{Incorporating external knowledge for meeting requirements}

Incorporating external knowledge in order to  manipulate  exploration behavior also plays a fundamental role in solving problems in which  the  speed of convergence is of secondary importance to operational requirements.   One line of research involves devising some mechanism (reward-related or otherwise) to force the agent to conform to certain operational specifications that are expressed in temporal logic 
\cite{bacchus1996rewarding, littman_environment-independent_2017,   brafman_ltlfldlf_2018,  camacho_ltl_2019,li_reinforcement_2017, icarte_using_2018, hasanbeig_logically-constrained_2019, giacomo_restraining_2020}.  
  Another is to cast the problems in the context of {\em safe reinforcement learning}, where the objective is to ensure that the agent achieves some acceptable level of return while meeting safety requirements \cite{garcia_comprehensive_2015,pecka_safe_2014}.
 Safety in this context is usually quantified by some risk measure or is characterized by some prescribed  behavior of the agent.
  Risk-based approaches for safe reinforcement learning use risk measures that are mostly derived from  (i) the variance of return and (ii) the probability of the agent entering an error state or visiting some undesirable state-action pair.
   For such approaches, external knowledge is incorporated for risk evaluation, while meeting safety requirements means minimizing the risk.
     
Two current main trends of risk-averse reinforcement learning   are (1) to modify the optimization criteria so as to include some form of risk measure, and (2) to use a risk measure directly for action selection. Techniques of the first trend include those that (i) use  a worst-case criterion, where a policy is considered to be optimal if it has the maximum worst-case return \cite{heger1994consideration,gaskett_reinforcement_2003,garcia_safe_2012}, (ii) use  risk-sensitive criteria  based on exponential functions  \cite{mihatsch_risk-sensitive_2002, geibel_risk-sensitive_2005}, and (iii) introduce safety-related constraints into the problem formulation so that solutions can be obtained under the framework of Constrained Markov Decision Processes 
\cite{geibel_reinforcement_2006, moldovan_safe_2012}. 
The second trend focuses on retaining the original (i.e., non-safety related)  optimization criteria while using a risk measure to directly influence action selection. This leads to a variety of techniques, including those that (i)  restrict exploration to a set of safe policies \cite{cheng_end--end_2019} or a set of states that are considered controllable \cite{gehring_smart_2013}, (ii) regulate the exploration via safety signals \cite{dalal_safe_2018}, and (iii) ensure that exploration is stable in the sense of Lyapunov (e.g., \cite{berkenkamp2018safe}).

Safety can also be characterized by certain prescribed  behavior of the agent, with the prescription expressed in symbolic logic. This is the formalism in the approach of {\em shielding}, where external knowledge is encapsulated in logic-based specifications that prescribe the safe behavior, and the agent is manipulated to meet these specifications during exploration
\cite{
alshiekh_safe_2017, 
tappler_automata_2022, konighofer_online_2022, hasanbeig_cautious_2020, bastani_safe_2021, den_hengst_planning_2022}. 

\subsection{Novelty,  key limitation, and application context}

Whether for improving speed of convergence or for meeting requirements on exploration behavior, essentially all the approaches cited in Section \ref{sec:literature} aim to guide the agent in avoiding  certain states or state-action pairs  that are considered undesirable or  unsafe,  under the premise that visiting them may result in a deterioration of expected return, violation of some requirement, or at the extreme a catastrophic outcome.   Since avoiding a specific state implies avoiding some action(s) at the predecessor of the specified state, state avoidance is subsumed under state-action avoidance, which is generally referred to as {\em action pruning} in the reinforcement learning literature. In these approaches, action pruning 
is {\em static} in the sense that, if an action is to be avoided by the agent at a state, it will be avoided at that state for all visits during the entire learning process. 
Such static action pruning breaks persistent exploration, giving rise to the possibility of finding only policies that are sub-optimal (i.e., when compared to an optimal policy obtainable without any safety considerations).

The problem of constrained exploration (with  optimality preservation) as proposed in this paper is also about manipulating the exploration behavior of the agent. 
The key technical difference between our proposed approach and those reported in the literature (and thereby the novelty) is that in our approach  action pruning is {\em dynamic} in the sense that the decision of whether to prevent the agent from taking a specific action at a state depends not on the state itself but on how that state is reached. In other words,  this decision depends on the sequence of actions that the agent has executed in order to reach that state. 
 With dynamic  action pruning, the agent may be prevented by the supervisor  from taking a specific action on one visit to a state but be permitted to take that action  on another visit to the same state, if these two visits are resulted from  two different sequences of actions (both starting from the initial state). Such supervision admits the possibility that  all state-action pairs are `visitable', thus enabling the agent to maintain persistent exploration
 even though its behavior is constrained by some specification throughout the learning process. In particular, the construction  of such a supervisor is independent of the dynamics of the environment.

Moreover,  the establishment  in this work of a necessary and sufficient condition (on the specification) for maintaining persistent exploration makes it possible to determine {\em a priori}, for the case where the environment dynamics is assumed to be  deterministic and known, whether optimality with respect to the unconstrained problem can still be achieved under constrained exploration.  Although there exist approaches (such as Constrained Markov Decision Process discussed in Section \ref{sec:literature}) for solving a constrained reinforcement-learning problem under the same assumption, these approaches do no directly address the question of whether optimality is preserved without actually solving the problem. 

While the implementation of constrained exploration (under a supervisor that performs dynamic action pruning) does not require any knowledge about the environment,  the method proposed in this work for determining whether optimality is preserved in such a setting does require the  dynamics of the environment to be deterministic and known. This requirement represents a key limitation on the applicability of the theoretical results reported in this paper, which is elaborated further in Section \ref{sec:discussion}.

Despite this limitation, the proposed approach remains useful  in certain application domains of reinforcement learning. One such domain is the Teacher-Student framework e.g., \cite{torrey2013teaching, zimmer2014teacher}. In this framework, the teacher is an expert who provides guidance to a student (i.e., the learning agent). The teacher typically has access to a complete model of the environment. 
Under this framework,  the teacher may attempt to guide the student by restricting the actions that the student is allowed to take at a given state.
With a deterministic environment,  the consequence of an action taken by the student is predictable, which enables the teacher to craft more effective guidance, e.g., for improving the speed of learning or ensuring the safety of the student. In addition, when establishing objectives for teaching or for the student's learning,  
it would be desirable that (before the student commences the learning process)  the teacher is able to determine whether optimality with respect to the unconstrained setting can still be achieved while the learning behavior of the student is constrained by some given specification. With the assumption that the dynamics of the environment is deterministic, our proposed approach can be directly applied to facilitate this determination.

This application context (in which learning is conducted in a known deterministic environment) can be extended to the setting that integrates Curriculum Learning into the Teacher-Student framework.
Under such a framework, the teacher can design a curriculum that gradually introduces the student to more complex tasks. For instance, a curriculum can include {\em learning tasks that transition from solving deterministic problems} to non-deterministic ones.
This  allows the student to focus on basic tasks at the early stage of learning without having to deal with the added complexity of uncertainty and randomness, thus helping the student learn more effectively.

\subsection{Related work}

In developing our proposed approach, we employ a general methodology for studying two interacting state-transition systems. In our problem setting, the specification is represented by the language generated by a state-transition model (in the form of an automaton),  and the dynamics of the agent when unconstrained is modeled by another a state-transition model  (again, an automaton). The exploration behavior of the agent that meets this specification is then investigated  in the context of the `synchronized' operation of these two state-transition models.

This general  methodology is also employed in a number of existing approaches  on the subject of manipulating the exploration behavior of the agent to satisfy given specifications. In these approaches, the specifications are first expressed in temporal logic then converted into some state-transition model, and the state-transition models representing the specifications and the dynamics of agent-environment interaction are in the form of either  automata or  Markov Decision Processes. Although the specific objectives of these existing approaches are different from that presented in this paper, in terms of  methodology  these existing approaches collectively represent the state-of-the-art to which our work is the most closely related.   Littman et al.\ 
 \cite{littman_environment-independent_2017}   
 and Brafman et al.\  \cite{brafman_ltlfldlf_2018}  studied the specficication  of non-Markovian rewards using linear temporal logic; 
De Giacomo et al.\  proposed the concept of {\em restraining bolts}  to  force the agent to exhibit certain prescribed behavior \cite{giacomo_restraining_2020}; Alshiekh et al.\ \cite{alshiekh_safe_2017} and den Hengst et al.\ \cite{den_hengst_planning_2022} proposed the technique of {\em shielded learning}, based on the concept of {\em shielding} in reactive systems \cite{konighofer2017shield}, to enforce safety specifications; Hasanbeig et al.\  proposed a method for limiting the extent of exploration to a portion that is relevant to satisfying some logic-based constraints so as to improve the speed of learning \cite{hasanbeig_logically-constrained_2019} and     developed the concept of {\em safe padding} to guide the agent in conforming to safety specifications \cite{hasanbeig_cautious_2020}. 
 In these approaches, an optimal policy is sought for the constrained reinforcement-learning problem resulted from the imposition of specifications on the behavior of the agent,  while in our proposed approach we still seek an optimal policy for the  {\em unconstrained} case (i.e., to preserve optimality) even though the behavior of the agent is also constrained to meet certain  specifications.
Such optimality preservation is made possible by the dynamic action-pruning mechanism of the supervisor that enables the agent to  maintain persistent exploration while meeting the specifications.

\subsection{Organization}

Section \ref{sec:prelim} summarizes notations and terminologies related to the existing concepts and techniques that are used in the subsequent analysis.  Section \ref{sec:automaton_model} presents the proposed automaton-model that describes the dynamics of the class of unconstrained reinforcement-learning systems investigated in this paper.  Section \ref{sec:supervized_rl} presents the structure and characteristics of supervision, and discusses automaton  representation of  desired behavior. Section \ref{sec:existence_optimality} proves the existence of an effective supervisor for a given desired behavior, and presents a necessary and sufficient condition for preserving optimality in a reinforcement-learning process under supervision. Section \ref{sec:discussion} discusses the implications, limitations, and possible extension of our approach. Section \ref{sec:conclusion} presents the conclusions.

\section{Preliminaries}
\label{sec:prelim}

The set of real numbers is denoted by $\mathbb{R}$, and the  set of non-negative integers by $\mathbb{N}_{0}$. A  closed interval between $a \in \mathbb{R}$ and $b \in \mathbb{R}$ is denoted by $[a, b]$, and an integer interval between $i \in \mathbb{N}_{0}$ and $j \in \mathbb{N}_{0}$ by $[\![ i, j ]\!]$. Let $|B|$ denote the number of elements in a finite set $B$, and $2^{B}$ denote the power set of $B$. Let $\varnothing$ denote the empty set.
 An alphabet is a nonempty finite set (denoted by $\mathcal{A}$)  whose member are symbols. A string over $\mathcal{A}$ is a sequence of symbols each belonging to $\mathcal{A}$.   In this paper, we use the terms {\em string} and {\em action sequence} interchangeably depending on the context. The concatenation of two
strings $u$ and $v$, written as  $uv$, is the string
consisting of $u$ followed by $v$.
 Let $|s|$ denote the
length of a string $s$, i.e., the number of symbols in $s$. 
Let $\lambda$ denote the empty string with the 
properties: (i) $|\lambda| = 0$, and (ii)  
$\lambda \, l  = l \, \lambda = l$
for any string $l$. Let $\mathcal{A}^{+}$ be the set of all possible strings over $\mathcal{A}$. Let $\mathcal{A}^{\ast} = \lambda \cup \mathcal{A}^{+}$. 
If $l = usv$ with $u, s, v \in \mathcal{A}^{\ast}$, then  we call $u$ a {\em prefix} of $l$ (written as $u \leq l$) and $s$ a {\em substring} of $l$ (written as $s \prec l$). A language over $\mathcal{A}$ is a subset of $\mathcal{A}^{\ast}$.  The prefix-closure of a language $L$, denoted by $\overline{L}$, is defined as: 
$\overline{L} = \{ u  \in \mathcal{A}^{\ast} \,|  u \leq l  \mbox{ for some } l \in L \}$.  A language $L$ is {\em prefix-closed} if $L = \overline{L}$.  (In this paper, we use $L$ as a generic symbol to denote a language  and use the calligraphic form $\mathcal{L}$ to denote a `specification language'.)

A deterministic automaton consists of six elements expressed as: 
\[
(\mathcal{S}, \mathcal{A}, \delta, \Gamma, s^{\circ}, \mathcal{S}^{\tiny \bullet})
\] 
where
$\mathcal{S}$ is a finite set of states, $\mathcal{A}$ is a finite set of events (i.e., an alphabet); $\delta : \mathcal{S} \times \mathcal{A} \rightarrow \mathcal{S}$ is called the transition function; $\Gamma: \mathcal{S} \rightarrow 2^{\mathcal{A}}$  is called the active event function; $s^{\circ} \in \mathcal{S}$ is the initial state; and 
$\mathcal{S}^{\bullet} \in \mathcal{S}$ is the set of marked states.  We say that a transition $\delta(s,a)$ is {\em defined}, written as $\delta(s, a)!$, if there is a state $s^{\prime} \in \mathcal{S}$ such that $\delta(s,a) = s^{\prime}$. The members of $\mathcal{A}$ are called {\em actions} in reinforcement learning literature; in this paper, we will follow this convention.  For a state $s$, $\Gamma(s)$ specifies the set of all actions $a$ for which $\delta(s,a)$ is defined. We  refer to $\Gamma(s)$ as the {\em active event set} at $s$. In a typical reinforcement-learning problem, $\Gamma(s)$ contains all the actions available for selection by the agent at state $s$, i.e., $\Gamma(s) = \mathcal{A}$.

The transition function $\delta$ can be extended
from $\mathcal{S} \times \mathcal{A}$ to $\mathcal{S} \times \mathcal{A}^{\ast}$ 
according to the following rules:
$\delta(s, \lambda)  =  s$ and 
$\delta(s, l a) = \delta(\delta(s, l), a)$,
where $s \in \mathcal{S}$, $l \in \mathcal{A}^{\ast}$, and $a \in \mathcal{A}$,
provided that $\delta(s, l)$ and $\delta(\delta(s, l),a)$ are defined. A state $s \in \mathcal{S}$ is {\em reachable} ({\em coreachable}) if there is a string $l \in \mathcal{A}^{\ast}$ such that $\delta(s^{\circ},l)=s$ (resp.\ $\delta(s, l) \in \mathcal{S}^{\bullet}$). 
An automaton is {\em trim} if all of its states are both reachable and coreachable.

We use a bold capital letter to represent a specific automaton, in which case each element of the automaton is assigned a subscript in the form of the lower case of that letter. For instance,
$	\mathbf{X}= (\mathcal{S}_{x}, \mathcal{A}_{x}, \delta_{x}, \Gamma_{x}, s^{\circ}_{x}, \mathcal{S}^{\bullet}_{x})$. Let
$\mathcal{R}_{e}[\mathbf{X}]$ denotes the reachable part of $\mathbf{X}$.
The language  {\em generated} by $\mathbf{X}$, denoted by $L(\mathbf{X})$,  is defined as:
$L(\mathbf{X}) := \{ l \in \mathcal{A}_{x}^{\ast} \,| \,\delta_{x}(s_{x}^{\circ}, l)! \}$.
The language {\em marked} (or {\em recognized}) by $\mathbf{X}$, denoted by $L_{m}(\mathbf{X})$, is defined as:
$L_{m}(\mathbf{X}) := \{ l \in \mathcal{A}_{x}^{\ast} \,|\, \delta_{x}(s_{x}^{\circ}, l) \in \mathcal{S}_{x}^{\bullet} \}$.  If $\mathbf{X}$ is trim, then $L(\mathbf{X}) = \overline{L_{m}(\mathbf{X})}$.  A language is {\em regular} if it is recognized by some finite automaton.
We say that two strings $u, v \in \mathcal{A}_{x}^{\ast}$ are equivalent  with respect to $\mathbf{X}$, written as $u \sim_{\mbox{\tiny $\mathbf{X}$}} \!v$, if there exists a state $s \in \mathcal{S}_{x}$ such that $\delta(s_{x}^{\circ}, u) = \delta(s_{x}^{\circ}, v) = s$.  The relation $\sim_{\mbox{\tiny $\mathbf{X}$}}$ divides $\mathcal{A}_{x}^{\ast}$ into equivalence classes, one for each state that is reachable from $s_{x}^{\circ}$ \cite{hopcroft1979introduction}.  
We denote the equivalent class corresponding to state $s$ as $\mathbb{C}_{x}(s)$.

The {\em product} of two automata $\mathbf{X}$ and $\mathbf{Y}$ is defined as \cite{cassandras2008introduction} 
\[
\mathbf{Z} := \mathbf{X} \|\, \mathbf{Y} = \mathcal{R}_{e}\big [ 
(\mathcal{S}_{z}, \mathcal{A}_{z}, \delta_{z},\Gamma_{z},  (s^{\circ}_{x}, s^{\circ}_{y}), \mathcal{S}^{\bullet}_{z})\big]
\]
where 
$\mathcal{S}_{z} = \mathcal{S}_{x} \times \mathcal{S}_{y}$,  $\mathcal{A}_{z} = \mathcal{A}_{x} \cup  \mathcal{A}_{y}$, 
$\mathcal{S}^{\bullet}_{z} = \mathcal{S}^{\bullet}_{x} \times\mathcal{S}^{\bullet}_{y}$, $\Gamma_{z} ( (s_{x}, s_{y})) = \Gamma_{x} (s_{x}) \cap \Gamma_{y}(s_{y})$ with  $(s_{x}, s_{y}) \in \mathcal{S}_{z}$, and the joint transition function $\delta_{z}$ is defined as
\[
\delta_{z} (( s_{x}, s_{y}),\, a) = \left \{
\begin{array}{ll}
	\big( \delta_{x} (s_{x},\, a),\, \delta_{y} (s_{y},\, a) \big) & \mbox{if} \,\, a \in \Gamma_{x}(s_{x}) \cap \Gamma_{y}(s_{y}) \\
	\mbox{undefined} & \mbox{otherwise} 
\end{array}
\right . 
\]
It follows from the definition of $\delta_{z}$ that $l \in \mathbb{C}_{z}( (s_{x}, s_{y}))$ implies that $l \in \mathbb{C}_{y}(s_{y}) \cap \mathbb{C}_{x}(s_{x})$.
 If $\mathcal{A}_{x} \subseteq \mathcal{A}_{y}$ and $L(\mathbf{X}) \subseteq L(\mathbf{Y})$, then  $\mathbf{X}$ is isomorphic to $\mathbf{X}\| \mathbf{Y}$ with state renaming. Hence, we have 
 $L(\mathbf{X}\|\mathbf{Y}) = L(\mathbf{X})$ and $L_{m}(\mathbf{X}\|\mathbf{Y}) = L_{m}(\mathbf{X})$.

We next introduce the notion of the probabilistic language generated by a stochastic system consisting of an automation with a state-wise action distribution. Detailed definition and analysis concerning such a system  have been presented 
in \cite{lawford1993supervisory,pantelic2009probabilistic,garg1999probabilistic}. We highlight here the key points relevant to analysis presented in this paper.
	Consider an automaton $\mathbf{X}$. At each state $s \in \mathcal{S}_{x}$, assign a discrete probability distribution $\pi_{b}$ over $\mathcal{A}_{x}$ such that 
$\pi_{b}(s, a) > 0$ if and only if $\delta_{x}(s, a)!$ for all $a \in \mathcal{A}_{x}$ and $\sum_{a \in \mathcal{A}_{x}} \pi_{b}(s, a) = 1$. 
  The probabilistic language generated by $(\mathbf{X}, \pi_{b})$, denoted by $L^{p}_{\mathbf{x}}(l)$, is defined as:
$L^{p}_{\mathbf{x}} : \mathcal{A}_{x}^{\ast} \rightarrow [0, 1]$, where $L^{p}_{\mathbf{x}}(\lambda) = 1$, and for all  $l \in L(\mathbf{X})$ and $a \in \mathcal{A}_{x}$,
\[
L^{p}_{\mathbf{x}}(la) = \left \{ \begin{array}{ll}
L^{p}_{\mathbf{x}}(l) \, \pi_{b} \left( 
\delta_{x}(s^{\circ}_{x}, l), a \right) & 
\mbox{if } \delta_{x}(s^{\circ}_{x}, l)! \\
L^{p}_{\mathbf{x}}(la) = 0 & \mbox{otherwise}
\end{array} \right.
\]
Informally, $L^{p}_{\mathbf{x}}(l)$ is the probability of the string $l$ being generated by $\mathbf{X}$.

\section{Automaton Model of Reinforcement Learning}
\label{sec:automaton_model}

Reinforcement-learning problems are typically formulated as Markov Decision Processes, in which the dynamics of the interaction between the agent and its environment is described by a state-transition structure. If we view the agent taking an action (when it is at a state) as  `generating' a symbol denoting  that action, then the sequences of actions thus generated by the agent (as it traverses through the states) can be considered to form a formal language that describes the exploration behavior of the agent. This perspective motivates the choice of formulating reinforcement-learning problems in the framework of automata and formal language theory, with the benefit that this framework offers a rich set of concepts and techniques for studying such dynamical behavior.

In this section we formulate the dynamics of an  reinforcement-learning process with unconstrained exploration in the framework of automata and formal language theory. 

\subsection{Unconstrained learning behavior}

We consider the class of episodic reinforcement-learning problems where an agent operates in an environment modeled by the trim automaton:
\begin{equation}
	\mathbf{G} = (\mathcal{S}_{g}, \mathcal{A}_{g}, \delta_{g}, \Gamma_{\!\!g}, s^{\circ}_{g},\mathcal{S}^{\bullet}_{g})
	\label{eq:RL_Automaton}
\end{equation}
with  the following properties:
(i)  $|\mathcal{S}_{g}| < \infty$; 
(ii) the transition function $\delta_{g}$ is {\em total}, i.e., $\delta_{g}(s,a)!$ for all $(s,a) \in \mathcal{S}_{g} \times \mathcal{A}_{g}$; and (iii)  $s^{\circ}_{g}$ and  $\mathcal{S}^{\bullet}_{g}$ are known.   We also call a state in $\mathcal{S}^{\bullet}_{g}$ a {\em goal} state. 
 Since $\delta_{g}$ is total, we have $\Gamma_g(s) = \mathcal{A}_{g}$ for all $s \in \mathcal{S}_{g}$.

\begin{figure}[!h]\centering
	\scalebox{0.9}[0.9]{\includegraphics{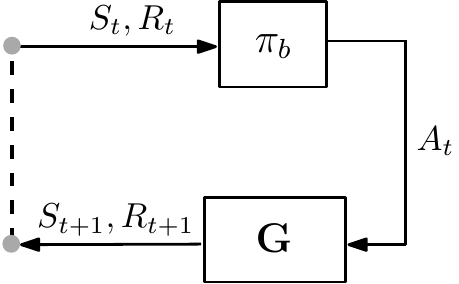}}
	\caption{Unconstrained learning process. The dashed line indicates the advance of time; that is, $(\cdot)_{t+1}$ in the previous step becomes  $(\cdot)_{t}$ in the current step.}
	\label{fig:unsupervised_RL}
\end{figure}

An episode starts at  $t=0$  with the agent at  $s^{\circ}_{g}$ and terminates when the agent reaches  a state in $\mathcal{S}^{\bullet}_{g}$. 
The interaction between the agent and its environment can be characterized by a feedback structure as illustrated in Figure \ref{fig:unsupervised_RL}.  Let $S_{t}$ and $A_{t}$ denote the state where the agent is in, and the action taken by the agent, at time step $t$, respectively.    
We say that the state-action pair $(S_{t}, A_{t})$ is {\em visited} by the agent when at  $S_{t}$ the agent executes  $A_{t}$ according to its behavior policy $\pi_{b}$. 

We do not need to be concerned with the exact form of $\pi_{b}$; we just require it to have the property (as discussed in Section \ref{sec:prelim}) that for all $(s, a) \in \mathcal{S}_{g} \times \Gamma_{g}(s)$, 
\begin{equation}
\pi_{b}(s, a) > 0
\label{eq:pi_b}
\end{equation}
 if and only if $\delta_{g}(s, a)!$. Typical behavior policies, such as  $\epsilon$-greedy and softmax, have this property.
 
Upon executing $A_{t}$ at time $t$ the agent reaches the next state $S_{t+1}$  as determined by $\delta_{g}$ at time $(t+1)$ and receives  a reward $R_{t+1} = \rho(S_{t}, A_{t})$, where   $\rho: \mathcal{S}_{g} \times \mathcal{A}_{g} \rightarrow \mathbb{R}$ is the reward function. The $Q$-value of a state-action pair $(s,a) \in \mathcal{S}_{g} \times \mathcal{A}_{g}$, defined in the standard way and denoted by $Q(s,a)$, is estimated iteratively in accordance with some pre-defined update rule  based on the rewards received \cite{sutton2018reinforcement, bertsekas2019reinforcement}.  
  We refer to $(\mathbf{G}, \pi_{b})$ as the  {\em unconstrained}  learning process.

\subsection{Optimality in an unconstrained learning process}
\label{sec:unsup}

In practice, during learning the agent will carry out a finite number of episodes until the state-action values $Q(s,a)$ are considered to have converged to their optimum values, denoted by $Q^{\ast}(s,a)$, based on which an optimal policy  can be obtained. 
Let $A^{(i)}_{t} \in \mathcal{A}_{g}$ denote the action {\em taken} by the agent at time step $t$ during episode $i$, and $l^{(i)}_{t}$ denote the sequence of actions taken by the agent up to and including time step $(t-1)$,  with $l^{(\cdot)}_{0} = \lambda$ (i.e., the empty string). Then we can write
$l^{(i)}_{t} = A^{(i)}_{0} A^{(i)}_{1} \cdots A^{(i)}_{t-1} \in \mathcal{A}^{\star}_{g}$.  By definition, $\big|l^{(i)}_{t}\big| = t$.  
Let $N \in \mathbb{N}_{0}$ be the total number of episodes in a learning process, and let $n_{i}$ denote the total number of time steps in  episode $i$. The language and the marked language generated by the agent over $N$ episodes are
$L = {\textstyle{\cup_{i=1}^{N}}} \overline{l}^{(i)}_{n_{i}}$ and $L_{m} = {\textstyle{\cup_{i=1}^{N}}}  l^{(i)}_{n_{i}}$,
respectively. For readability we sometimes omit the episode index $(i)$, e.g., by writing  $l_{t}$ instead of $l^{(i)}_{t}$. 
 With $\pi_{b}(s,a) >0$ for all $(s,a) \in (\mathcal{S}_{g} - \mathcal{S}_{g}^{\bullet}) \times \mathcal{A}_{g}$, we have $L \rightarrow L(\mathbf{G})$ and $L_{m} \rightarrow L_{m}(\mathbf{G})$ as $N \rightarrow \infty$. 
We refer to  $L(\mathbf{G})$ and $L_{m}(\mathbf{G})$ as the language and the marked language  generated by $(\mathbf{G}, \pi_{b})$, respectively,
under the assumption that in practice a very large number of episodes  are conducted.

With a properly designed update rule, the $Q$-values in an unconstrained learning system $(\mathbf{G}, \pi_{b})$ will converge to their optimums under the Robbins-Monro conditions.  
To satisfy these conditions requires all state-action pairs be visited infinitely often by the agent during the learning process.  In the context of this paper, which assumes pre-existence of a properly designed $Q$-value update rule,  achieving optimality means meeting this requirement.

\begin{proposition} \label{prop:VisitAll} 
Consider an unconstrained learning process $(\mathbf{G}, \pi_{b})$. In any episode every state-action pairs $(s,a) \in (\mathcal{S}_{g}-\mathcal{S}_{g} ^{\bullet}) \times \mathcal{A}_{g}$ has a non-zero probability of being visited by the agent.
 \end{proposition}

\noindent{\bf  Proof}: 
For a state-action pair  $(s, a)$ to be visited by the agent in an unconstrained learning process $(\mathbf{G}, \pi_{b})$, the following two conditions must be met. First, the agent must (starting from $s^{\circ}_{g}$) reach the state $s$ by executing some string $l \in \mathbb{C}_{g}(s)$. Second, the action $a$ must be selected by the behavior policy $\pi_{b}$ when the agent is at state $s$. Using the notation of probability languages presented in Section \ref{sec:prelim}, the probability of a string $l \in \mathbb{C}_{g}(s)$ being generated can be expressed as $L^{p}_{\mathbf{G}}(l)$. Let $p_{\pi}(a|l)$ denote the probability that  action $a$ is taken by the agent at state $s$ after the agent has executed the string $l$ to reach $s$ starting from $s^{\circ}_{g}$, and let  $p_{u}(s,a)$ denote the probability of a state-action pair  $(s, a)$ being visited by the agent. Then we can express this probability as
 \begin{equation}
 	p_{u}(s,a) = p_{\pi} (a|l) \cdot L^{p}_{\mathbf{G}}(l) 
 \end{equation}
We note that, by definition, $p_{\pi} (a|l) = \pi_{b} (s,a)$.

Since $\mathbf{G}$ is reachable and $\pi_{b}(s,a)>0$ for all $(s,a) \in (\mathcal{S}_{g}-\mathcal{S}_{g} ^{\bullet}) \times \Gamma_{g}(s)$, there exists a string $l$ such that $l \in \mathbb{C}_{g}(s)$ for any $s \in \mathcal{S}_{g}$. 
In an episode $i$, the probability of a string $l^{(i)}_{t} \in \mathcal{C}_{g}(s)$ 
being generated is $L^{p}_{\mathbf{G}}(l^{(i)}_{t})$. 
With $A_{t} = a$, $S_{t} = s$, and noting that $l^{(\cdot)}_{0}$ is the empty string (i.e.,  $l^{(\cdot)}_{0} = \lambda$), we have \begin{eqnarray}
p_{u}(s,a)  & \equiv & 
p_{u}(S_{t}, A_{t}) \nonumber \\
& = & 
p_{\pi}(A_{t}|l^{(i)}_{t}) \cdot L^{p}_{\mathbf{G}}(l^{(i)}_{t}) \nonumber \\
& = & 
p_{\pi}(A_{t}|l^{(i)}_{t}) \cdot p_{\pi}(A_{t-1}|l^{(i)}_{t-1}) \cdot L^{p}_{\mathbf{G}}(l^{(i)}_{t-1}) \nonumber \\
& \vdots & \mbox{\small (Continuing expansion till  $L^{p}_{\mathbf{G}}(l^{(i)}_{0})$ is reached)}\nonumber \\
& = & \underbrace{p_{\pi}(A_{t-1} | l_{t-1}) \cdots p_{\pi}(A_{0} | l^{(i)}_{0})}_{\mbox{\footnotesize $t$ terms}}
\cdot \underbrace{
L^{p}_{\mathbf{G}}(l_{0})
}_{
L^{p}_{\mathbf{G}}
(\lambda)=1
} \nonumber \\
& = &  \prod^{t-1}_{k=0} p_{\pi}(A_{k} | l_{k}) 
= \prod^{t-1}_{k=0} \pi_{b}(S_{k}, A_{k})
> 0
\label{eq:L_p_G}
\end{eqnarray}
Hence, in any episode the probability of any steta-action pair $(s,a) \in (\mathcal{S}_{g}-\mathcal{S}_{g} ^{\bullet}) \times \mathcal{A}_{g}$ being visited by the agent is non-zero. 
 \hfill $\Box$
 
As the number of episodes goes to infinity, every state-action pairs in $(s,a) \in (\mathcal{S}_{g}-\mathcal{S}_{g} ^{\bullet}) \times \mathcal{A}_{g}$ will being visited by the agent infinitely often.

\section{Constrained Learning Behavior}
\label{sec:supervized_rl}

To manipulate the exploration behavior of the agent is to impose on  $(\mathbf{G}, \pi_{b})$ an external mechanism that prevents (when necessary)  the agent from  taking certain action at a state, such that the language thus generated by the agent is restricted to a subset of $L(\mathbf{G})$. 
 We call this mechanism a {\em supervisor}, denoted by $\Pi$, and refer to $(\mathbf{G}, \pi_{b}, \Pi)$ as a   {\em constrained} learning process. Any mention of $\mathbf{G}$ hereafter is with reference to  $(\mathbf{G}, \pi_{b})$ or $(\mathbf{G}, \pi_{b}, \Pi)$ depending on the context, unless it is stated otherwise.
  
 We refer to the set of actions available to the agent at a state as the {\em admissible action set}.  
 In $(\mathbf{G}, \pi_{b})$, the admissible action set at a state $s$ is  $\Gamma_{g}(s)$.  In  $(\mathbf{G}, \pi_{b}, \Pi)$, the supervisor may prevent the agent from taking a certain action at a state $s = \delta_{g}(s^{\circ}, l)$ by removing it from $\Gamma_{g}(s)$, thus reducing the admissible action set to a subset of $\Gamma_{g}(s)$. The supervisor decides which action (if any) to remove based on $l$.  An action is said to be {\em disabled} if it is removed by the supervisor, and is said to be {\em enabled} otherwise. Only enabled actions belong to the admissible action set, which is denoted by $\Pi(l)$. Hence,  $\Pi(l) \subseteq \Gamma_{g}(s) = \mathcal{A}_{g}$, and
  $\pi_{b} (s,a) = 0$ for all $a \notin \Pi(l)$.

 \begin{definition}
 A supervisor $\Pi$  is a function from the language $L(\mathbf{G})$ to the power set of $\mathcal{A}_{g}$, i.e.,
$\Pi: L(\mathbf{G}) \rightarrow 2^{\mathcal{A}_{g}}$.
 \end{definition}
 
 	By definition a supervisor is any mechanism that can disable actions at a state. For example, a mechanism that randomly disables actions at a state is a supervisor. In Section \ref{sec:existence_optimality}, we will discuss the form of a supervisor (referred to as an ``effective supervisor'') that carries out such action-disablement selectively so that the agent under supervision will generate a specific language.

\begin{remark}
The concept of supervision by action-disablement and the realization of supervisor as automaton discussed in this section and in Section \ref{sec:ExistenceRealization}, respectviely, are adapted from the Supervisory Control Theory that focuses on the control of discrete-event systems \cite{ramadge1987supervisory}. Central to  this  adaptation is the establishment of an extended feedback-control structure (involving $\mathbf{G}$, $\pi_{b}$ and $\Pi$) in the context of the class of reinforcement learning problems under investigation. 
\label{remark:SCT}
\end{remark}

\subsection{Structure and characterization of supervision}
\label{sec:supervision}

\begin{figure}[!h]\centering
\scalebox{0.9}[0.9]{\includegraphics{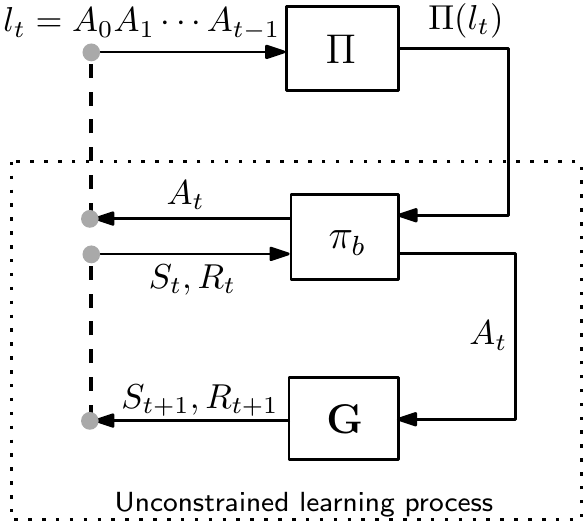}}
\caption{Feedback structure of a constrained learning process.  A dashed line indicates the advance of time; that is, $(\cdot)_{t+1}$ in the previous step becomes  $(\cdot)_{t}$ in the current step.} 
\label{fig:BlockDiagram}
\end{figure}

The interaction between $\Pi$ and $(\mathbf{G}, \pi_{b})$ in a constrained learning process is illustrated in Figure \ref{fig:BlockDiagram}. 
Suppose that the agent, starting from the initial state $S_{0} = s^{\circ}_{g}$ at time step $t=0$  and following some action sequence $l_{t} \in L(\mathbf{G})$, reaches the state $S_{t}$ at time step $t>0$, i.e., $S_{t}  = \delta_{g}(S_{0}, l_{t})$. At time step $t$ 
the supervisor $\Pi$ generates (based on $l_{t}$) the admissible action set $\Pi(l_{t})$, from which the behavior policy $\pi_{b}$ selects one action (if $\Pi(l_{t})$ is non-empty) that the agent is to execute. Upon taking an action $A_{t}$ at time step $t$, the agent reaches the next state $S_{t+1}$ and receives a reward $R_{t+1}$. The supervisor then generates, with $l_{t+1} = l_{t}\, A_{t}$, the next admissible action set $\Pi(l_{t+1})$  from which $\pi_{b}$ selects the next action. This cycle repeats until the episode terminates. 

  Under the structure as illustrated in Figure \ref{fig:BlockDiagram}, the supervisor $\Pi$ exerts a dynamic feedback control on $(\mathbf{G}, \pi_{b})$. It is {\em dynamic} because the decision on whether to disable an action  at a state $s = \delta_{g}(s^{\circ}_{g}, l)$ depends not on $s$ itself but on how $s$ is reached; that is, for two different action sequences $l_{1}, l_{2} \in \mathbb{C}_{g}(s)$,  $\Pi$ may disable an action $a$ when $s$ is reached  via $l_{1}$ but may not do so when $s$ is reached via $l_{2} \neq l_{1}$.

\begin{definition}
The  language generated by the agent under the supervision of $\Pi$, denoted by $L(\Pi/\mathbf{G})$,  is defined recursively as follows:
\begin{itemize}
	\item [(1)] $\lambda   \in  L(\Pi/\mathbf{G})$ 
	\item[(2)] $\Big[\, l \in L(\Pi/\mathbf{G}) \,\,\, \mbox{and} \,\,\,  l a \in L(\mathbf{G}) \,\,\, \mbox{and} \,\,\,  a \in \Pi(l)\, \Big] \Leftrightarrow l a \in L(\Pi/\mathbf{G})$
		\end{itemize}
\end{definition}

By definition, $L(\Pi/\mathbf{G})$ is prefix-closed, i.e., $L(\Pi/\mathbf{G}) = \overline{L(\Pi/\mathbf{G})}$. 
If $\Pi(l) \subset \Gamma_{g}(s)$ for some $s \in \left(\mathcal{S}_{g} - \mathcal{S}^{\bullet}_{g} \right)$, then the effect of $\Pi$ is to restrict the behavior of the agent to some sublanguage of $L(\mathbf{G})$, i.e., $L(\Pi/\mathbf{G}) \subset L(\mathbf{G})$. 

\subsection{Specification language}
\label{sec:spec}

We consider only desired behavior that can be expressed as a regular and prefix-closed sublanguage of $L(\mathbf{G})$. We call such a language a {\em specification language} and denote it by $\mathcal{L}$, i.e., $\mathcal{L} \subset L(\mathbf{G})$.
Since $\mathcal{L}$ is regular,
we can construct a finite state automaton, denoted by $\mathbf{H}$, that recognizes $\overline{\mathcal{L}}$ (i.e., the prefix-closure of $\mathcal{L}$); that is  $L(\mathbf{H}) = \overline{\mathcal{L}}$.  (In fact, $\mathbf{H}$ also recognizes $\mathcal{L}$ since $\mathcal{L}$ is assumed to be prefixed-closed, i.e., $\mathcal{L} = \overline{\mathcal{L}}$.) We will show in Section \ref{sec:existence_optimality} that under the supervision of $\Pi$ the behavior of the agent, i.e., the language $L(\Pi/\mathbf{G})$, is the same as the desired behavior, i.e., $L(\Pi/\mathbf{G}) = L(\mathbf{H}) = \mathcal{L}$.

In practice, we can first express the desired agent  behavior  in natural language, then encode the logical relationships implied in that behavior in the transition structure of  $\mathbf{H}$. Methods of constructing $\mathbf{H}$  for some typical logical relationships are discussed in \cite{cassandras2008introduction}. Such an automaton plays a key role in the realization of an effective supervisor, as discussed in  Section \ref{sec:ExistenceRealization}.

\begin{example} \label{exp:1}
Consider an unconstrained learning process $(\mathbf{G}_{1}, \pi_{b1})$ in which the agent is to traverse from an initial state to a goal state in a two-dimensional grid-world. At a state the agent can take one of the four actions to move {\em up} (denoted by $a_{1}$), {\em right} ($a_{2}$), {\em down} ($a_{3}$), and {\em left} ($a_{4}$), resulting in the agent reaching deterministically the adjacent state in the direction of the intended action, or remaining in the same state if it collides with a border of the grid. Suppose that it is desirable to prevent the agent from oscillating between two adjacent (non-bordered) states by taking the action sequence $a_{1}a_{3}$ or $a_{2}a_{4}$.   The specification language for this desired behavior consists of all strings in $L(\mathbf{G}_{1})$ except those that contain $a_{1}a_{3}$ or $a_{2}a_{4}$ as a substring, i.e., $\mathcal{L}_{1} = L(\mathbf{G}_{1}) - \mathcal{L}^{\prime}_{1}$, where $
\mathcal{L}^{\prime}_{1} := \{ l \in L(\mathbf{G}_{1}) \, \big |\, a_{1}a_{3} \prec l \,\,\, \mbox{or} \,\,\, a_{2}a_{4}  \prec l \}$.
An automaton that generates $\mathcal{L}_{1}$ is shown in Figure \ref{fig:SpecificationExample}. By construction,  $L(\mathbf{H}_{1}) = \mathcal{L}_{1}$ is regular and prefix-closed. 
\begin{figure}[!h]\centering
	\scalebox{0.6}[0.6]{\includegraphics{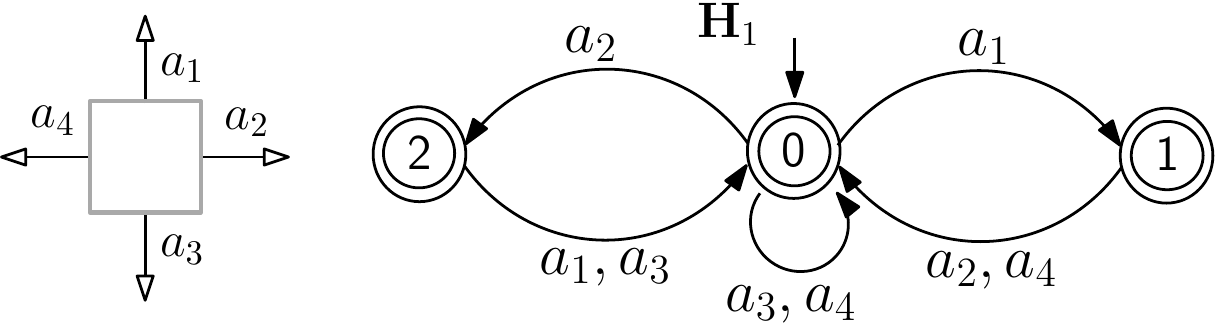}}
	\caption{Automaton $\mathbf{H}_{1}$ that generates the language $\mathcal{L}_{1}$. The entering arrow indicates the initial state; a double circle indicates a marked state.}
	\label{fig:SpecificationExample}
\end{figure}
\end{example}

\section{Effectiveness of Supervisor and Optimality Preservation}
\label{sec:existence_optimality}

\begin{definition} Consider an unconstrained learning process $(\mathbf{G}, \pi_{b})$ and a  prefix-closed language $\mathcal{L} \subset L(\mathbf{G})$. A supervisor $\Pi$ is {\em effective} with respect to $\mathcal{L}$ if $L(\Pi/\mathbf{G}) = \overline{\mathcal{L}}$. 
\end{definition}

In this section, we prove  that, given a specification language $\mathcal{L} \subset L(\mathbf{G})$, an effective supervisor exists. We then establish a necessary and sufficient condition under which an effective supervisor is optimality-preserving.

\subsection{Existence and realization  of  supervisor}
\label{sec:ExistenceRealization}

\begin{lemma} \label{lemma:Pi existence} 
For a given unconstrained learning process $(\mathbf{G}, \pi_{b})$ and a  prefix-closed language $\mathcal{L} \subset L(\mathbf{G})$,  an effective supervisor exists. 
\end{lemma}

\noindent
{\bf  Proof}: Since $\mathcal{L}$ is prefix-closed, we have $\mathcal{L} = \overline{\mathcal{L}}$.
Define a supervisor
\begin{equation}
\Pi(l) = \big \{a \in \mathcal{A}_{g} \,|\, l \in L(\mathbf{G}) \,\,\mbox{and}\,\, la \in \overline{\mathcal{L}} \big \}
\label{eq:Pi}
\end{equation} 
We will carry out the proof by induction on the length of strings of $L(\Pi/\mathbf{G})$ and $\overline{\mathcal{L}}$ to show that, with $\Pi$ as defined above,  $l \in L(\Pi/\mathbf{G})$ if and only if $l \in \overline{\mathcal{L}}$.  The base case of $l = \lambda$ is trivial since $\lambda$ belongs to both languages by definition. The hypothesis is that 
$l  \in L(\Pi/\mathbf{G})$ if and only if $l \in \overline{\mathcal{L}}$.   The inductive step involves strings of the form $l a$, with $a \in \mathcal{A}_{g}$.  For the {\sf IF} case, we have $l a \in \overline{\mathcal{L}}$.  Since $\overline{\mathcal{L}}$ is prefix-closed, we have $l \in \overline{\mathcal{L}}$; consequently, $l \in L(\Pi/\mathbf{G})$ (from the hypothesis). Since $l \in  L(\Pi/\mathbf{G}) \subset L(\mathbf{G})$ and $l a \in \overline{\mathcal{L}}$, from Equation (\ref{eq:Pi}) we have  $a \in \Pi(l)$. Hence,  $la \in L(\Pi/\mathbf{G})$. 
For the {\sf ONLY-IF} case, $la \in L(\Pi/\mathbf{G})$ implies that  $l \in L(\Pi/\mathbf{G})$  (since $L(\Pi/\mathbf{G})$ is prefix-closed), which by the hypothesis implies that $l \in \overline{\mathcal{L}}$. Now $l \in L(\Pi/\mathbf{G})$ and $la \in L(\Pi/\mathbf{G})$ implies that $a \in \Pi(l)$. Therefore, $la \in \overline{\mathcal{L}}$. \hfill $\Box$

The function $\Pi$ can be represented as an automaton that `reads'  a string $l$ and generates the admissible action set $\Pi(l)$. When operating in the structure illustrated in Figure \ref{fig:BlockDiagram}, the interaction between $\mathbf{H}$ and $(\mathbf{G}, \pi_{b})$ can be implemented via the product $\mathbf{H}\| \mathbf{G}$;  that is, the language generated by the agent under the supervision of $\Pi$, i.e., $L(\Pi/\mathbf{G})$, is the same as the language generated by $\mathbf{H}\| \mathbf{G}$ \cite{wonham2019supervisory, cassandras2008introduction}. Specifically, the automaton $\mathbf{H}$ is said to be a {\em realization} of  $\Pi$ if $L(\mathbf{H}\| \mathbf{G}) = L(\Pi/\mathbf{G})
$ and $L_{m}(\mathbf{H}\| \mathbf{G})   = L_{m}(\Pi/\mathbf{G})$.

\begin{remark} 
If during an episode  $\Pi (l_{t})$ is empty at time step $t$ for a string $l_{t}\in C_g\left(s\right)\cap L\left(\Pi/\mathbf{G} \right)$, then no action at the state $s$ is enabled by the supervisor. Consequently, no action is available for selection by the behavior policy $\pi_{b}$ at $s$ at time step $t$. If $s$ is not a marked (i.e., goal) state, then the agent will deadlock at $s$,  and the episode ends prematurely at time step $t$. The behavior of the agent up to time step $t$ still meets the specification $\mathcal{L}$.
\end{remark}

\begin{proposition} A trim automaton $\mathbf{H}$ (with  $\mathcal{A}_{h} = \mathcal{A}_{g}$ and  all states marked) is a realization of the supervisor $\Pi$ as defined in Equation (\ref{eq:Pi})  if $L(\mathbf{H}) = \mathcal{L}$, where  $\mathcal{L} = \overline{\mathcal{L}} \subset L(\mathbf{G})$.
\label{prop:realization}
\end{proposition} 

\noindent
{\bf  Proof}:  Since $\mathbf{H}$ is trim and $L(\mathbf{H}) = \mathcal{L}$ is prefix-closed, we have  $L_{m}(\mathbf{H}) = L(\mathbf{H}) = \overline{\mathcal{L}}$.
Now $L(\mathbf{H}\|\mathbf{G}) = L(\mathbf{H}) \cap L(\mathbf{G}) = \overline{\mathcal{L}} \cap L(\mathbf{G})$. Since $\Pi$ is effective, we have  $L(\mathbf{H}\|\mathbf{G}) = \overline{\mathcal{L}}  = L(\Pi/\mathbf{G})$. 
 For the marked language $
L_{m}(\mathbf{H}\| \mathbf{G})$, we have $
L_{m}(\mathbf{H}\| \mathbf{G}) = L_{m}(\mathbf{H}) \cap L_{m}(\mathbf{G}) = \overline{\mathcal{L}} \cap L_{m}(\mathbf{G}) = L(\Pi/\mathbf{G}) \cap L_{m}(\mathbf{G}) = L_{m}(\Pi/\mathbf{G})$.
 The last step  reflects the fact that, because all states of $\mathbf{H}$ are marked,  marking in $(\mathbf{G}, \pi_{b}, \Pi)$ is done  based on $\mathcal{S}^{\bullet}_{g}$  and does not involve $\Pi$.  \hfill $\Box$

\begin{remark} The proofs of Lemma \ref{lemma:Pi existence} and Proposition \ref{prop:realization} imply that, given a prefix-closed language $\mathcal{L} = L(\mathbf{H}) \subset L(\mathbf{G})$, an effective supervisor can be realized using the automaton $\mathbf{H}$ without any explicit knowledge of the transition structure of $\mathbf{G}$.  
\label{remark:Pi_and_G}
\end{remark}

\subsection{Optimality preservation}
\label{sec:optprv}

The existence of an effective supervisor (established in Lemma \ref{lemma:Pi existence}) naturally gives rise to the question of whether, with its behavior restricted by $\Pi$ to $\overline{\mathcal{L}} \subset L(\mathbf{G})$, the agent is still able to find the intrinsic optimums of the $Q$-values; that is, whether optimality in $(\mathbf{G}, \pi_{b})$ is preserved under the supervision of  $\Pi$. Recall that a requirement for satisfying the Robbins-Monro conditions in  $(\mathbf{G}, \pi_{b})$, as discussed in Section \ref{sec:unsup}, is that all state-action pairs be visited infinitely often;  that is, the probability of any action in $\mathcal{A}_{g}$ being taken by the agent at any state in $(\mathcal{S}_{g}-\mathcal{S}^{\bullet}_{g})$ must be non-zero.  Optimality preservation  means that this requirement is satisfied in $(\mathbf{G}, \pi_{b}, \Pi)$.

For a state-action pair  $(s, a)$ to be visited by the agent in  $(\mathbf{G}, \pi_{b}, \Pi)$, the following three conditions must be met. First, the agent must (starting from $s^{\circ}_{g}$) reach the state $s$ by executing some string $l \in \mathbb{C}_{g}(s) \cap L(\Pi/\mathbf{G})$. Second, the action $a$ must be enabled by the supervisor $\Pi$ at $s$. Third, the action $a$ must be selected by the behavior policy $\pi_{b}$ when the agent is at state $s$. 
 Suppose that a supervisor $\Pi$ is realized by an automaton $\mathbf{H}$. Then we have  $L(\Pi/\mathbf{G}) = L(\mathbf{H}\| \mathbf{G})$.   
Recall from the proof of Proposition \ref{prop:VisitAll} that $p_{\pi}(a|l)$ denotes the probability that the action $a$ is taken by the agent after the agent has executed the action sequence $l$ to reach $s$ starting from $s^{\circ}_{g}$. 
Now let $p_{\mbox{\tiny $\Pi$}}(a|l)$ denote the probability that the action $a$ is enabled by $\Pi$ after the agent has executed $l$ to reach $s$ starting from $s^{\circ}_{g}$. Using the notation of probability languages presented in Section \ref{sec:prelim}, the probability of a string $l \in \mathbb{C}_{g}(s) \cap L(\Pi/\mathbf{G})$ being generated by the agent under the supervision of $\Pi$ can be expressed as $L^{p}_{\mbox{\tiny $\mathbf{H}\|\mathbf{G}$}}(l)$. Then the probability of a state-action pair $(s, a)$ being visited by the agent under the supervision of $\Pi$, denoted by $p(a|s)$, can be expressed as 
\begin{equation}
p(a|s) = p_{\pi}(a|l) \cdot p_{\mbox{\tiny $\Pi$}}(a|l) 
\cdot 
	L^{p}_{\mbox{\tiny $\mathbf{H}\|\mathbf{G}$}}(l)
\label{eq:prob_state_action_pair}
\end{equation}

\begin{lemma}
\label{lemma:L_p_epsilon} $ L^{p}_{\mbox{\tiny $\mathbf{H}\|\mathbf{G}$}}(l) >0$. 
\end{lemma}

 This lemma is needed in the proof of the main result stated in Theorem \ref{theorem:main}.
 The poof for Lemma \ref{lemma:L_p_epsilon} is presented in  Appendix A; it follows the same line of argument based on the expansion of $l$ (in terms of its individual constituent actions taken by the agent at each time step) as shown in the proof of Proposition \ref{prop:VisitAll}.

\begin{definition} \label{def:optimality_preserving}
A supervisor $\Pi$  in a constrained learning process $(\mathbf{G}, \pi_{b}, \Pi)$ is said to be {\em optimality-preserving}  if $p(a|s)>0$ for all  $(s,a) \in (\mathcal{S}_{g} - \mathcal{S}^{\bullet}_{g}) \times \mathcal{A}_{g}$.
\end{definition}

By definition a supervisor can be optimality-preserving without being effective. A  trivial example is a supervisor that does not disable any action at any state.
 In the context of this paper, whether an effective supervisor is optimality-preserving  depends on the  desired behavior, as is illustrated in Example \ref{exp:visitability} in Section \ref{sec:the_problem}. 

We next establish a condition for a given effective supervisor to be optimality-preserving in  terms of a property of the specification language $\mathcal{L}$. For this purpose, we introduce the notion of an automaton being `covered' by some sublanguage of the language generated by that automaton. 

\begin{definition} \label{def:cover}
Let $\mathcal{L} \subset L(\mathbf{G})$.   We say that $\mathcal{L}$ {\em covers} $\mathbf{G}$  if, for any $(s, a) \in (\mathcal{S}_{g} - \mathcal{S}^{\bullet}_{g}) \times \mathcal{A}_{g}$, there exists a string $l \in \mathbb{C}_{g}(s) \cap \mathcal{L}$ such that $l a \in \mathcal{L}$. 
\end{definition}

Informally, a language $\mathcal{L}$ covering an automaton $\mathbf{G}$ means that for  any non-marked state $s$ in $\mathbf{G}$ and  any action $a \in \mathcal{A}_{g}$, we can always find a string $l$ in $\mathcal{L}$ satisfying the following two properties: (1) starting from the initial state $s^{\circ}_{g}$ the state $s$ can be reached by following $l$, and 
(2) the string $la$ also belongs to $\mathcal{L}$.
 It is possible that, of all the strings that have property 1, some do not have property 2. As long as for each state-action  pair there is at least one string in the specification language  that satisfies both properties, then the coverage holds. In the context of a constrained learning process, this notion of coverage means that, in order to satisfy the specification, the supervisor may block an action at a state if that state is reached via some action sequence but will not do so if the same state is reached via a {\em different} action sequence.
 This notion of coverage, in conjunction with the main result presented below, will be illustrated in two scenarios in Section \ref{sec:Example}. 

\color{black}
We now present the main result.

\begin{theorem} \label{theorem:main}
An effective  supervisor  $\Pi$  is optimality-preserving if and only if $\mathcal{L}$ covers $\mathbf{G}$. 
\end{theorem}

\noindent
{\bf  Proof}: Since $\Pi$ is effective, we have $L(\Pi/\mathbf{G}) = \mathcal{L}$. We first prove the {\sf IF} case.  
Since $\mathcal{L}$ covers $\mathbf{G}$, there exists a string $l \in \mathbb{C}_{g}(s) \cap \mathcal{L} = \mathbb{C}_{g}(s) \cap L(\Pi/\mathbf{G}) $ such that $l a \in  L(\Pi/\mathbf{G})$ for any $(s, a) \in (\mathcal{S}_{g} - \mathcal{S}^{\bullet}_{g}) \times \mathcal{A}_{g}$.
Now  from Equation (\ref{eq:prob_state_action_pair}), we have
$p(a |s) = p_{\pi}(a | l) \cdot p_{\mbox{\tiny  $\Pi$}}(a | \, l) \cdot L^{p}_{\mbox{\tiny $\mathbf{H}\|\mathbf{G}$}}(l)$.  
Since $l \in  L(\Pi/\mathbf{G})$ and $l a \in  L(\Pi/\mathbf{G})$, we have $a \in \Pi(l)$, i.e., $a$ is enabled by $\Pi$ at $s$. So $p_{\mbox{\tiny  $\Pi$}}(a | \, l) = 1$ and  $\pi_{b}(s,a)>0$. 
By definition $p_{\pi} (a | l)= \pi_{b}(s, a)$; so $p_{\pi} (a | l) >0$.  From Lemma \ref{lemma:L_p_epsilon}, we have $L^{p}_{\mbox{\tiny $\mathbf{H}\|\mathbf{G}$}}(l) >0$. 
Hence, $p(a|s) >0$.

We next prove the {\sf  ONY-IF} case. Specifically, we will show that, under an effective $\Pi$ if $p(a |s) > 0$ for any $(s, a) \in (\mathcal{S}_{g} - \mathcal{S}^{\bullet}_{g}) \times \mathcal{A}_{g}$, then there exists a string $l \in \mathbb{C}_{g}(s) \cap L(\Pi/\mathbf{G})$ such that  $l a  \in L(\Pi/\mathbf{G})$.
Consider any  $(s, a) \in (\mathcal{S}_{g} - \mathcal{S}^{\bullet}_{g}) \times \mathcal{A}_{g}$ with some $l \in \mathbb{C}_{g}(s)$. From Equation (\ref{eq:prob_state_action_pair}),  $p(a |s)>0$ 
implies  that there exists a string $l \in \mathbb{C}_{g}(s)$ such that $p_{\pi}(a|l)>0$, $p_{\mbox{\tiny $\Pi$}}(a |\, l)>0$, 
and  $L^{p}_{\mbox{\tiny $\mathbf{H}\|\mathbf{G}$}}(l) > 0$. 
From Equation (\ref{eq:Lemma2}) in the appendix,  we have $L^{p}_{\mbox{\tiny $\mathbf{H}\|\mathbf{G}$}}(l)  \equiv L^{p}_{\mbox{\tiny $\mathbf{H}\|\mathbf{G}$}}(l_{t})  =  \prod^{t-1}_{k=0} \, p_{\pi}(A_{k} | l_{k}) \cdot p_{\mbox{\tiny  $\Pi$}}(A_{k} | l_{k})$.
So $L^{p}_{\mbox{\tiny $\mathbf{H}\|\mathbf{G}$}}(l) >0$ implies that $p_{\mbox{\tiny $\Pi$}}(A_{k} | \, l_{k})>0$ (which in turn implies that $p_{\mbox{\tiny $\Pi$}}(A_{k} | \, l_{k})=1$) for all $k\in [\![\,0, t-1 ]\!]$. Hence, $A_{k} \in \Pi(l_{k})$ for all $k\in [\![\,0, t-1 ]\!]$. Consequently, we have $l_{t} \equiv l \in L(\Pi/\mathbf{G})$.  We next show that $la \in L(\Pi/\mathbf{G})$. Since $p_{\mbox{\tiny $\Pi$}}(a | l)>0$ implies that $p_{\mbox{\tiny  $\Pi$}}(a | l)=1$, we have $a \in \Pi(l)$. Therefore, $la \in L(\Pi/\mathbf{G}) = \mathcal{L}$.  \hfill $\Box$

Intuitively, if we think of the agent's movement through the state space as being described by a sequence of state-action pairs, e.g., $S_{0}A_{0}S_{1}A_{1}\cdots S_{t}A_{t}$, then an optimality-preserving supervisor will permit at least one trajectory to pass through any state-action pair (where the state is not a marked state). The existence of such permitted trajectories depends on  whether the specification covers $\mathbf{G}$.

\subsection{Deciding whether $\mathcal{L} \subset \mathbf{G}$ covers $\mathbf{G}$}
\label{sec:decide_cover}

We next present a sufficient condition  for deciding whether a given $\mathcal{L} \subset L(\mathbf{G})$ covers  $\mathbf{G}$.  This condition is based on the product of $\mathbf{G}$ and an automaton $\mathbf{H}$ that generates  $\mathcal{L}$, i.e., $L(\mathbf{H}) = \mathcal{L}$, and involves the notion of the visitablity of a state in $\mathbf{G}$.
Let $\mathbf{M} =   \mathbf{H} \| \mathbf{G}$. 
 Define a function $\Omega_{m} : \mathcal{S}_{g} \rightarrow 2^{\mathcal{S}_{m}}$   (with $\mathcal{S}_{m} = \mathcal{S}_{h} \times \mathcal{S}_{g}$) that maps a state $s_{g} \in \mathcal{S}_{g}$ to a set that contains just the states  in $\mathbf{M}$ whose second  component  is $s_{g}$, i.e., $\Omega_{m}(s_{g}) :=  \{ (a,b) \,|\, a \in \mathcal{S}_{h}, b \in \mathcal{S}_{g}, \,\, \mbox{and} \,\, b = s_{g} \}$. 
Consider in $\mathbf{M}$ two states $(s_{h}, s_{g})$ and $(s^{\prime}_{h}, s_{g})$
reached by two different strings  $l$ and $l^{\prime}$ that both belong to $\mathbb{C}_{g}(s_{g})$. By definition, 
 $(s_{h}, s_{g})$ and $(s^{\prime}_{h}, s_{g})$
belong to $\Omega_{m}(s_{g})$. It is possible that an action $a \in \mathcal{A}_{g}$ is disabled by $\Pi$ at  $(s_{h}, s_{g})$ but enabled at  $(s^{\prime}_{h}, s_{g})$.
 If after a number of visits by the agent to the states in $\Omega(s_{g})$ all actions in $\mathcal{A}_{g}$ have been enabled (at least once) by $\Pi$, 
then  all state-action pairs $(s_{g}, a)|_{a\in \mathcal{A}_{g}}$ have a non-zero probability of being visited by the agent. We characterize this property  as the {\em visitability} of a state.

\begin{definition} \label{def:visitable}
A state $s_{g} \in (\mathcal{S}_{g} - \mathcal{S}^{\bullet}_{g})$ in $\mathbf{G}$ is {\em visitable} 
with respect to $\mathbf{M}$ if 
\begin{equation}
\bigcup_{(s_{h}, s_{g})  \in \Omega_{m}(s_{g})}\Gamma_{m}((s_{h}, s_{g})) =\mathcal{A}_{g}
\label{eq:AllActive}
\end{equation} 
\end{definition}

\begin{lemma} 
If a state $s_{g} \in (\mathcal{S}_{g} - \mathcal{S}^{\bullet}_{g})$ is visitable with respect to $\mathbf{M}$, then for any $a \in \mathcal{A}_{g}$ there exits a state 
$ (s_{h}, s_{g}) \in \Omega_{m}(s_{g})$ such that $a \in \Gamma_{m}((s_{h}, s_{g}))$.
\label{lemma:Active_a_g}
\end{lemma}

\noindent
{\bf  Proof}: This lemma follows directly from Definition \ref{def:visitable}.  Suppose that there are $n$ states in $\mathbf{M}$ corresponding to $s_{g}$, i.e., 

\[
\Omega_{m}(s_{g}) = \{ (s_{h_{1}}, s_{g}), (s_{h_{2}}, s_{g}), \ldots, (s_{h_{n}}, s_{g}) \}
\]
Equation (\ref{eq:AllActive}) can then be expressed as 
\begin{equation}
\bigcup\limits_{(s_{h}, s_{g}) \in \Omega_{m}(s_{g})} \Gamma_{m}((s_{h}, s_{g})) = \Gamma_{m}((s_{h_{1}}, s_{g})) \cup   
\cdots \cup \Gamma_{m}((s_{h_{n}}, s_{g})) =  \mathcal{A}_{g} 
\label{eq:AllActive_2}
\end{equation}
Any $a \in \mathcal{A}_{g}$  must belong to at least one of the active action sets $\Gamma_{m}(\cdot)$ in Equation (\ref{eq:AllActive_2}). 
\hfill $\Box$

\begin{definition} 
An automaton $\mathbf{G}$ is visitable with respect to $\mathbf{M}$ if every state in $(\mathcal{S}_{g} - \mathcal{S}^{\bullet}_{g})$ is visitable with respect to $\mathbf{M}$.
\end{definition}

We now state a condition for deciding whether a given $\mathcal{L} \subset L(\mathbf{G})$ covers  $\mathbf{G}$. 

\begin{theorem} \label{theorem:check_cover}
If $\mathbf{G}$ is visitable with respect to $\mathbf{M}$,
then $\mathcal{L}$ covers $\mathbf{G}$.
\end{theorem}

\noindent{\bf  Proof}: Consider any state-action pair $(s_{g}, a) \in (\mathcal{S}_{g}- \mathcal{S}^{\bullet}_{g}) \times \mathcal{A}_{g}$.
From Lemma \ref{lemma:Active_a_g}, for any $a \in \mathcal{A}_{g}$ we have  $a \in \Gamma_{m}((s_{h}, s_{g}))$ for some $(s_{h}, s_{g}) \in \Omega_{m}(s_{g})$.  Since $\mathbf{M}$ is reachable by construction,  there exists a string  $l \in \mathbb{C}_{m}((s_{h}, s_{g}))$. Consequently, $l a \in L(\mathbf{M}) = L(\mathbf{H}) =  \mathcal{L}$.
Since  $(s_{h}, s_{g}) \in \Omega_{m}(s_{g})$, so $l \in \mathbb{C}_{m}((s_{h}, s_{g}))$ implies that $l \in \mathbb{C}_{g}(s_{g})$.  Therefore, we have $l \in \mathbb{C}_{g}(s_{g}) \cap \mathcal{L}$. Since $l \in \mathbb{C}_{g}(s_{g}) \cap \mathcal{L}$ and  $l a \in \mathcal{L}$ for any $(s_{g}, a) \in (\mathcal{S}_{g}- \mathcal{S}^{\bullet}_{g}) \times \mathcal{A}_{g}$, we conclude that  $\mathcal{L}$ covers $\mathbf{G}$.
\hfill $\Box$

Theorem \ref{theorem:check_cover} and Lemma \ref{lemma:Active_a_g}  provide a method for checking whether a given $\mathcal{L}$ covers $\mathbf{G}$:
 For each $s_{g} \in (\mathcal{S}_{g} - \mathcal{S}^{\bullet}_{g})$, determine $\Omega_{m}(s_{g})$. If Equation (\ref{eq:AllActive}) is satisfied for all $s_{g} \in (\mathcal{S}_{g} - \mathcal{S}^{\bullet}_{g})$, then $\mathcal{L}$ covers $\mathbf{G}$.

\subsection{Illustrative Scenarios}
\label{sec:Example}

In this section we illustrate the concept of a sublanguage of an automaton covering that automaton. We use  
 two scenarios  involving  a $4 \times 4$ grid environment $\mathbf{G}$ with the transition characteristics as described in Example \ref{exp:1}. The agent is to find an optimal policy  for it to reach the marked state at the lower right corner of the grid, starting from the initial state at the top left corner.  The first scenario concerns a specification language that covers $\mathbf{G}$; the second concerns one that does not.

\begin{figure}[!h]\centering
\scalebox{0.6}[0.6]{\includegraphics{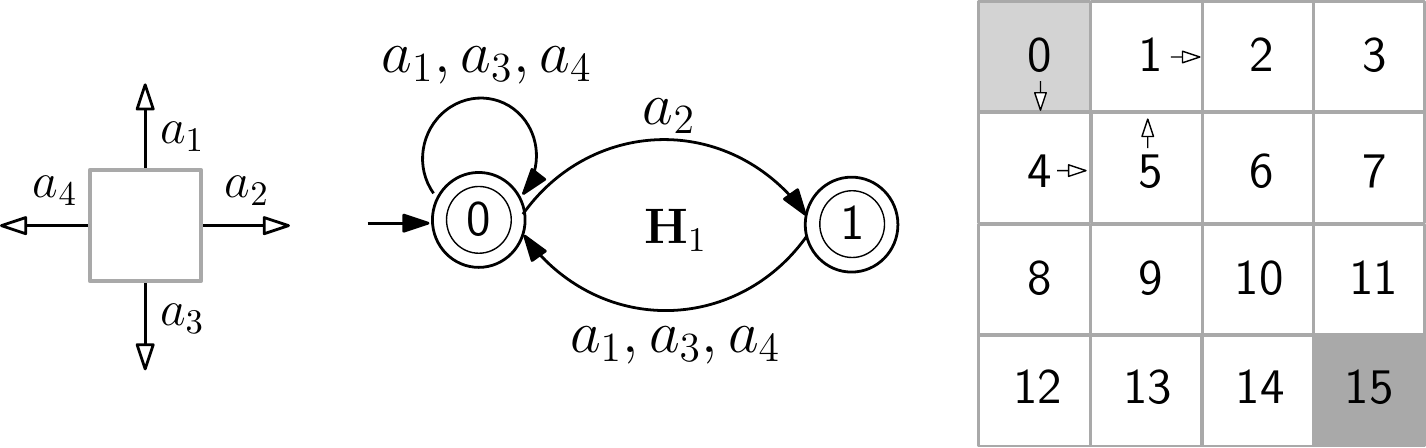}}
\caption{{\em Left}: Actions at a state. {\em Middle}: Automaton $\mathbf{H}_{1}$ representing the requirement that the agent is prohibited from taking two consecutive $a_{2}$ actions. {\em Right}: A $4 \times 4$ grid environment, with the initial state at {\sf 0} and the marked  state at {\sf 15}.   The four small arrows in the grid represent the action sequence $a_{3}\, a_{2}\, a_{1} \, a_{2}$.} 
\label{fig:ConstraintExample}
\end{figure}

In the first scenario, the agent is prohibited from taking two {\sf right} (i.e., $a_{2}$) actions consecutively. An automaton $\mathbf{H}_{1}$ that generates a prefix-closed specification language $\mathcal{L}_{1} = L(\mathbf{H}_{1}) \subset L(\mathbf{G})$  satisfying this requirement is shown in Figure \ref{fig:ConstraintExample}.

It can be checked that $\mathbf{G}$ is visitable with respect to $\mathbf{M}_{1}=\mathbf{H}_{1} \| \mathbf{G}$. 
Due to the nature of the specification language $\mathcal{L}_{1}$, only the  state-action pairs involving states that are not next to the left border and involving action $a_{2}$ need to be checked\,---\,since the transition structure in $\mathbf{G}$ does not allow a state next to the left border to be reached by an $a_{2}$ action.
Take the state-action pair $(1, a_{2})$ for instance. Although the agent is not allowed to take $a_{2}$ at state {\sf 1} after taking $a_{2}$ at state {\sf 0} (i.e., $a_{2}a_{2}$  cannot be a substring of any string in $\mathcal{L}_{1}$), it can still take action $a_{2}$ at state {\sf 1} via  the sequence $a_{3}\, a_{2}\, a_{1} \, a_{2}$ (which is a substring of a string in $\mathcal{L}_{1}$), as is illustrated by the small arrows in the grid in Figure \ref{fig:ConstraintExample}. So the state-action pair $(1,a_{2})$ in $\mathbf{G}$ is visitable with respect to $\mathbf{M}_{1}$. This can be verified by examining the transition structure of $\mathbf{M}_{1}$. Table \ref{tbl:ConstraintExample} shows the actions that are active at the two composite states involving state {\sf 1} of $\mathbf{G}$, namely, $(0,1)$ and $(1,1)$. Although $a_{2}$ is not active at $(1,1)$, it is active at $(0,1)$. Therefore, state {\sf 1} of $\mathbf{G}$ is visitable with respect to $\mathbf{M}_{1}$. The same analysis can be done for all the states in $\mathbf{G}$ that are not next to the left border to ascertain that  $\mathbf{G}$ is visitable with respect to $\mathbf{M}_{1}$.  So $\mathcal{L}_{1}$ covers $\mathbf{G}$.  Consequently, an effective supervisor  as defined in Equation (\ref{eq:Pi}) with respect to $\mathcal{L}_{1}$ is optimality-preserving.
 
\begin{table}[h!]
\centering
\caption{Active actions in $\mathbf{M}_{1}$  involving state {\sf 1} of $\mathbf{G}$.}
\begin{tabular}{cc|cc|cc} \hline
\multicolumn{2}{c|}{$\mathbf{H}_{1}$} & \multicolumn{2}{c|}{$\mathbf{G}$}  & \multicolumn{2}{c}{$\mathbf{M}_{1} =  \mathbf{H}_{1} \|\, \mathbf{G}$} \\   \hline 
$s_{h_{1}}$ & $a_{h_{1}}$ & $s_{g}$ & $a_{g}$ &  $s_{m_{1}}$ & $a_{m_{1}}$ \\ [0.03in] \hline
$0$ & $a_{1}$ & $1$ & $a_{1}$ & $(0,1)$ & $a_{1}$ \\ 
$0$ & $a_{2}$ & $1$ & $a_{2}$ & $(0,1)$ & $a_{2}$ \\ 
$0$ & $a_{3}$ & $1$ & $a_{3}$ & $(0,1)$ & $a_{3}$ \\ 
$0$ & $a_{4}$ & $1$ & $a_{4}$ & $(0,1)$ & $a_{4}$ \\ \hline
$1$ & $a_{1}$ & $1$ & $a_{1}$ & $(1,1)$ & $a_{1}$ \\ 
$1$ & $-$ & $1$ & $a_{2}$ & $(1,1)$ & $-$ \\ 
$1$ & $a_{3}$ & $1$ & $a_{3}$ & $(1,1)$ & $a_{3}$ \\
$1$ & $a_{4}$ & $1$ & $a_{4}$ & $(1,1)$ & $a_{4}$ \\ \hline
\end{tabular}
\label{tbl:ConstraintExample}
\end{table}

\begin{figure}[!h]
\centering\scalebox{0.45}[0.45]{\includegraphics[width=\textwidth]{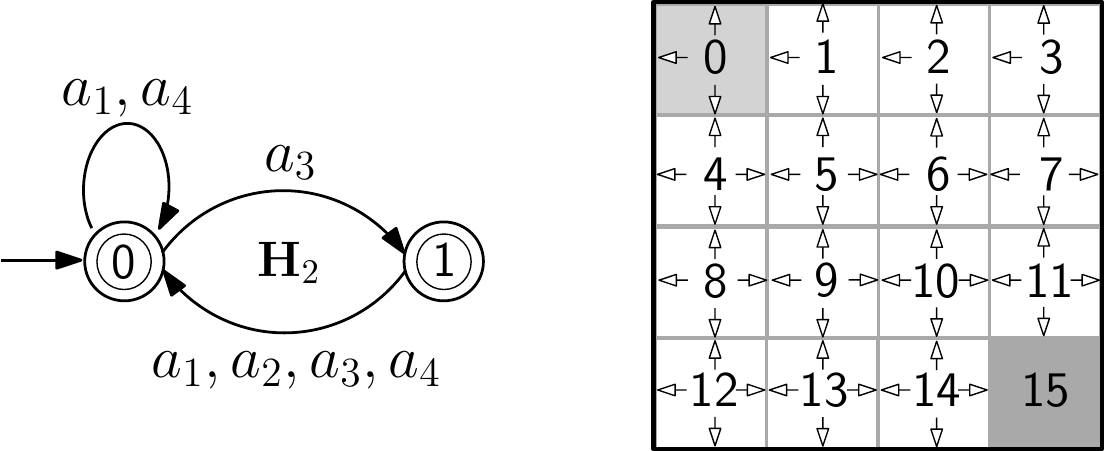}}
\caption{{\em Left}: Automaton $\mathbf{H}_{2}$ representing the requirement that the agent can take action $a_{2}$ only immediately after having taken $a_{3}$. {\em Right}: Behavior of the agent that satisfies $\mathcal{L}_{2}$.}
\label{fig:ConstraintNegExample}
\end{figure}

In the second scenario, the requirement is that the agent can take an $a_{2}$ action {\em only immediately} after having taken an $a_{3}$ action. An automaton $\mathbf{H}_{2}$ that generates a prefix-closed specification language $\mathcal{L}_{2} = L(\mathbf{H}_{2}) \subset L(\mathbf{G})$  satisfying this requirement is shown in Figure \ref{fig:ConstraintNegExample}. 

Since there is no downward action (i.e., $a_{3}$) that brings the agent to states {\sf 0}, {\sf 1}, {\sf 2}, and {\sf 3} in $\mathbf{G}$, we can conclude by inspection that to comply with the requirement the agent must not move to the right at those states, as is indicated by the missing right arrow at those states in Figure \ref{fig:ConstraintNegExample}. This conclusion can be verified by examining the transition structure of $\mathbf{M}_{2} = \mathbf{H}_{2} \| \mathbf{G}$.  We consider the state-action pair $(1,a_{2})$ in $\mathbf{G}$ as an illustration. In $\mathcal{S}_{h_{2}} \times \mathcal{S}_{g}$ the only two $(s_{h_{2}}$, $s_{g})$ pairs that involve state {\sf 1} of $\mathbf{G}$ are $(0,1)$ and $(1,1)$.   However, the pair $(1,1)$ is not a state in $\mathbf{M}_{2}$  because the only way for $\mathbf{H}_{2}$ to reach its own state {\sf 1}  is via an $a_{3}$  action but, due to the structure of $\mathbf{G}$, state {\sf 1} in $\mathbf{G}$ cannot be reached by an $a_{3}$ action. Consequently, of these two pairs only $(0,1)$ is a state in  $\mathbf{M}_{2}$. Table \ref{tbl:ConstraintNegExample} shows the actions that are active at $(0,1)$. Since $a_{2}$ is not active at state {\sf 0} of $\mathbf{H}_{2}$, it cannot be active at $(0,1)$. Therefore, state {\sf 1} of $\mathbf{G}$ is not visitable with respect to $\mathbf{M}_{2}$. Similar arguments can be made concerning the state-action pairs $(0,a_{2})$, $(2,a_{2})$, and $(3,a_{2})$ of $\mathbf{G}$. We thus conclude that $\mathcal{L}_{2}$ does not cover $\mathbf{G}$.
\begin{table}[h!]
\centering
\caption{Active actions in $\mathbf{M}_{2}$ involving state {\sf 1} of $\mathbf{G}$.}
\begin{tabular}{cc|cc|cc} \hline
\multicolumn{2}{c|}{$\mathbf{H}_{2}$} & \multicolumn{2}{c|}{$\mathbf{G}$}  & \multicolumn{2}{c}{$\mathbf{M}_{2} =  \mathbf{H}_{2} \| \mathbf{G}$} \\   \hline 
$s_{h_{2}}$ & $a_{h_{2}}$ & $s_{g}$ & $a_{g}$ &  $s_{m_{2}}$ & $a_{m_{2}}$ \\ [0.03in] \hline
$0$ & $a_{1}$ & $1$ & $a_{1}$ & $(0,1)$ & $a_{1}$ \\ 
$0$ & $-$ & $1$ & $a_{2}$ & $(0,1)$ & $-$ \\ 
$0$ & $a_{3}$ & $1$ & $a_{3}$ & $(0,1)$ & $a_{3}$ \\ 
$0$ & $a_{4}$ & $1$ & $a_{4}$ & $(0,1)$ & $a_{4}$ \\ \hline
\end{tabular}
\label{tbl:ConstraintNegExample}
\end{table}

\section{Discussion}
\label{sec:discussion}

In this section we first discuss the implications and limitations of our proposed approach. We then elaborate on a possible direction to extend the current work in the context of the theory of supervisory control of discrete-event systems.

The existence and realization of an effective and optimality-preserving supervisor, established in Lemma \ref{lemma:Pi existence} and Theorem \ref{theorem:main}, suggest a rigorous approach for manipulating the behavior of the agent during a reinforcement-learning process.  Under this approach,  achieving desired agent behavior is decoupled from achieving convergence of the learning process. 
Consequently, the choice of an update rule for the $Q$-values  
is independent of the design of a supervisor that restricts the agent  to certain prescribed behavior.
This decoupling makes possible the application of this approach in various scenarios that share an underlying theme of  manipulating the behavior of the agent.

The applicability of our proposed approach is limited by two main factors. 
The first concerns the assumption that a model of the environment $\mathbf{G}$ is known. Ideally we seek a supervisor that is {\em both} effective and optimality-preserving, so that we can manipulate the behavior of the agent without interfering with the determination of the intrinsic optimums; moreover, we would like to be able to ascertain the existence of such a supervisor  {\em a priori}. This is possible when a deterministic environment  model $\mathbf{G}$ is known, as shown in the proofs of Lemma \ref{lemma:Pi existence} and Theorem \ref{theorem:main}.   
In the  model-free  setting where $\mathbf{G}$ is unknown, even though we are still able to construct an effective supervisor in the form of an automaton $\mathbf{H}$ that generates $\mathcal{L}$ (see Remark \ref{remark:Pi_and_G}), the method presented in Section \ref{sec:optprv} is not applicable in checking {\em a priori} whether this effective supervisor is optimality-preserving, because doing so involves the computation of $\mathbf{H}\|\mathbf{G}$. A possible way of relaxing this assumption is to apply the framework of stochastic automata and probabilistic languages (e.g., \cite{garg1999probabilistic,kumar_control_2001}, so as to obtain some result concerning optimality preservation in the probabilistic sense.

The second limiting factor concerns the nature of supervision. In this paper the supervisor is assumed to be omnipotent in the sense that whenever it decides to disable an action $a \in \mathcal{A}_{g}$ at a state $s \in \mathcal{S}_{g}$, the behavior policy $\pi_{b}$ always complies with that decision, i.e., $\pi_{b}(s,a) = 0$. A more interesting control-theoretic issue arises if such compliance is not total with respect to $\mathcal{A}_{g}$;  that is, the supervisor is able to disable some but not all actions in $\mathcal{A}_{g}$. The question of how to deal with this  issue is addressed in the Supervisory Control Theory \cite{ramadge1987supervisory, wonham2019supervisory, cassandras2008introduction}, and thus serves as a motivation to extend the current work in the context of that theory.

In the Theory of Supervisory Control, an action is considered {\em controllable} if it can be disabled by the supervisor, and {\em uncontrollable} otherwise. The set of events (i.e., actions in this paper), denoted by $\Sigma$, is partitioned into two disjoint subsets $\Sigma_{c}$ and $\Sigma_{u}$, with $\Sigma_{c}$ containing events that are  controllable and $\Sigma_{u}$ uncontrollable. Under this arrangement, the aim of supervisory control is to ensure, {\em without} disabling any uncontrollable events, that the  behavior of the system under supervisory control stays within a specification language.
The scope of supervision as proposed in this paper in the context of reinforcement learning (in particular Lemma \ref{lemma:Pi existence}) covers a special case of that arrangement, where  all actions in $\mathcal{A}_{g}$  are assumed to be controllable, i.e., $\Sigma = \Sigma_{c} = \mathcal{A}_{g}$.
A particularly meaningful way to extend  the scope of supervision is to remove this assumption, so that pertinent control-theoretic properties (such as the  controllability of the behavior of the agent) can be  investigated. 
This extension would represent an important step in  establishing a formal connection between the Supervisory Control Theory and a class of reinforcement-learning systems.

\section{Conclusion}
\label{sec:conclusion}

We have introduced  the concept of constrained exploration in reinforcement learning with optimality preservation in terms of manipulating the behavior of the agent to meet some specification during the learning process. 
 We have investigated this concept in the context of the theory of automata and formal languages  by (i) establishing a feedback-control structure that models the dynamics of the unconstrained learning process,
(ii)  extending this structure by adding a supervisor to ensure that the behavior of the agent meets if possible a given specification, and (iii) establishing a necessary and sufficient condition (on the specification language) for optimality to be preserved under supervision.

The  work reported in this paper has  resulted in a formal method for manipulating the behavior of the agent in a class of reinforcement-learning systems.  
It demonstrates the utility and the prospect of studying reinforcement learning in the context of the theories of discrete-event systems, automata and formal languages.

\appendix
\section*{Appedix A: Proof of Lemma \ref{lemma:L_p_epsilon}}
\label{app:lemma_L_HG}
\noindent

 In Lemma \ref{lemma:L_p_epsilon}, the claim is that, for a state $s$ in a constrained learning system $(\mathbf{G}, \pi_{b}, \Pi)$, we have
$ L^{p}_{\mbox{\tiny $\mathbf{H}\|\mathbf{G}$}}(l) >0$ with  $l \in \mathbb{C}_{g}(s) \cap L(\Pi/\mathbf{G})$. The proof below follows the same line of argument based on the expansion of $l$ as shown in the proof of Proposition \ref{prop:VisitAll}. 
Recall that $|l| = t$. We first express $L^{p}_{\mbox{\tiny $\mathbf{H}\|\mathbf{G}$}}(l)$ in terms of the constituent actions of $l$ up to and including time step $(t-1)$, i.e.,  $l \equiv l_{t} = A_{0} A_{1} \cdots A_{t-1}$. Now,
\begin{eqnarray}
 L^{p}_{\mbox{\tiny $\mathbf{H}\|\mathbf{G}$}}(l) & \equiv &  L^{p}_{\mbox{\tiny $\mathbf{H}\|\mathbf{G}$}}(l_{t}) \nonumber \\
& = &  p_{\pi} (A_{t-1} |\, l_{t-1}) \cdot  
p_{\mbox{\tiny $\Pi$}}
(A_{t-1} |\, l_{t-1}) \cdot
 L^{p}_{\mbox{\tiny $\mathbf{H}\|\mathbf{G}$}}(l_{t-1})  \nonumber \\
& = & p_{\pi}(A_{t-1} | l_{t-1})  \cdot 
p_{\mbox{\tiny $\Pi$}}(A_{t-1} |\, l_{t-1}) \cdot 
p_{\pi}(A_{t-1} | l_{t-2})  \cdot p_{\mbox{\tiny  $\Pi$}}(A_{t-1} |\, l_{t-2}) \cdot  L^{p}_{\mbox{\tiny $\mathbf{H}\|\mathbf{G}$}}(l_{t-2})
\nonumber \\
& \vdots & \mbox{\small (Continuing expansion till  $L^{p}_{\mathbf{H}\|\mathbf{G}}(l_{0})$ is reached)}\nonumber \\
& = & \underbrace{p_{\pi}(A_{t-1} | l_{t-1}) \cdots p_{\pi}(A_{0} | l_{0})}_{\mbox{\footnotesize $t$ terms}}
\cdot 	 \underbrace{p_{\mbox{\tiny  $\Pi$}}(A_{t-1} | l_{t-1}) \cdots p_{\mbox{\tiny  $\Pi$}}(A_{0} | l_{0})}_{\mbox{\footnotesize $t$ terms}} \cdot \underbrace{
 L^{p}_{\mbox{\tiny $\mathbf{H}\|\mathbf{G}$}}(l_{0})
}_{
L^{p}_{\mbox{\tiny $\mathbf{H}\|\mathbf{G}$}
}
(\lambda)=1
} \nonumber \\
& = &  \prod^{t-1}_{k=0} p_{\pi}(A_{k} | l_{k}) \cdot p_{\mbox{\scriptsize  $\Pi$}}(A_{k} | l_{k})
\label{eq:Lemma2}
\end{eqnarray}
 Since $l \in L(\Pi/\mathbf{G})$, we have $p_{\mbox{\tiny  $\Pi$}}(A_{k} | l_{k}) = 1$; that is, $A_{k} \in  \Pi(l_{k})$ for all $k\in [\![\,0, t-1]\!]$.
 By definition $p_{\pi}(A_{k} | l_{k}) = \pi_{b}(S_{k}, A_{k})$, where $S_{k} = \delta_{g}(s^{\circ}_{g}, l_{k})$.
 Since $A_{k} \in  \Pi(l_{k})$ and $\pi_{b}(S_{k},A_{k}) >0$ if $A_{k} \in  \Pi(l_{k})$, we have $p_{\pi}(A_{k} | l_{k})  >0$ for all $k\in [\![\,0, t-1]\!]$. Therefore, $ L^{p}_{\mbox{\tiny $\mathbf{H}\|\mathbf{G}$}}(l_{t}) >0$. \hfill $\Box$


\end{document}